\documentclass[letterpaper]{article} 
\usepackage{aaai24}  
\usepackage{times}  
\usepackage{helvet}  
\usepackage{courier}  
\usepackage[hyphens]{url}  
\usepackage{graphicx} 
\usepackage{natbib}  
\usepackage{caption} 
\usepackage{algorithm}
\usepackage{algorithmic}

\usepackage{newfloat}
\usepackage{listings}
\DeclareCaptionStyle{ruled}{labelfont=normalfont,labelsep=colon,strut=off} 
\floatstyle{ruled}
\newfloat{listing}{tb}{lst}{}
\floatname{listing}{Listing}
\graphicspath{{figures/}}

\title{Constrained Meta-Reinforcement Learning for Adaptable Safety Guarantee \\with Differentiable Convex Programming}
\author {
    Minjae Cho\textsuperscript{\rm 1},
    Chuangchuang Sun\textsuperscript{\rm 2}
}
\affiliations {
    \textsuperscript{\rm 1}Department of Mechanical Engineering, Mississippi State University, MS 39762\\
    \textsuperscript{\rm 2}Department of Aerospace Engineering, Mississippi State University, MS 39762\\
    mc3216@msstate.edu, csun@ae.msstate.edu 
}

\usepackage{amsfonts, adjustbox, makecell}
\usepackage{amsmath}
\DeclareMathOperator*{\argmax}{argmax}
\DeclareMathOperator*{\argmin}{argmin}

\newcommand{\mD}{\mathcal{D}}
\newcommand{\mT}{\mathcal{T}}
\usepackage{mathtools}
\usepackage{cleveref}

\usepackage{color}

\begin{document}

\maketitle

\begin{abstract}

Despite remarkable achievements in artificial intelligence, the deployability of learning-enabled systems in high-stakes real-world environments still faces persistent challenges. For example, in safety-critical domains like autonomous driving, robotic manipulation, and healthcare, it is crucial not only to achieve high performance but also to comply with given constraints. Furthermore, adaptability becomes paramount in non-stationary domains, where environmental parameters are subject to change. While safety and adaptability are recognized as key qualities for the new generation of AI, current approaches have not demonstrated effective adaptable performance in constrained settings. Hence, this paper breaks new ground by studying the unique challenges of ensuring safety in \textit{non-stationary} environments by solving constrained problems through the lens of the meta-learning approach (learning-to-learn). While unconstrained meta-learning already encounters complexities in end-to-end differentiation of the loss due to the bi-level nature, its constrained counterpart introduces an additional layer of difficulty, since the constraints imposed on task-level updates complicate the differentiation process. To address the issue, we first employ successive convex-constrained policy updates across multiple tasks with differentiable convex programming, which allows meta-learning in constrained scenarios by enabling end-to-end differentiation. This approach empowers the agent to rapidly adapt to new tasks under non-stationarity while ensuring compliance with safety constraints. We also provide a theoretical analysis demonstrating guaranteed monotonic improvement of our approach, justifying our algorithmic designs. Extensive simulations across diverse environments provide empirical validation with significant improvement over established benchmarks.

\end{abstract}

\section{Introduction}

Artificial intelligence (AI) has made significant progress in the past few decades, ranging from mastering board games (AlphaGo~\cite{silver2016mastering}), and predicting protein structure (AlphaFold2~\cite{jumper2021highly}) to generating human-like texts (GPT-4~\cite{brown2020language}). Though it has the potential to revolutionize human society like electricity did about one hundred years ago, currently, its real-world impact in high-stakes scenarios is still yet proven beyond games. It is observed that, despite those significant successes, deployable Learning-Enabled Systems (LES, \cite{marcus2019rebooting}) are far less pervasive. One of the most critical concerns among others towards deployability is safety. In many scenarios, the safety of LES is not compromisable, especially those with humans in the loop. For example, autonomous driving vehicles should guarantee the safety of the drivers and other entities by following the driving rules and operating under various internal and external disturbances, such as partial sensor dysfunction and weather conditions. In healthcare and medicine, the treatment should guarantee that the side effects should not exceed the prescribed threshold. Therefore, learning-enable components should rigorously guarantee safety, and failing to do so can result in undesirable or even disastrous outcomes.


In this paper, we focus on the safety issues of reinforcement learning (RL), a popular framework for sequential decision-making. 
A significant amount of effort has been made to advance safe RL.
For example, constrained reinforcement learning~\cite{chow2017risk, achiam2017constrained} offers a compelling solution for training policy safely and responsibly by complying with safety constraints. 
However, there are still major gaps toward deployability in more restrictive environmental assumptions. Consider a \textit{non-stationary} environment with dynamically changing specifications, including the safety criterion. In this case, safety-aware RL policies trained point-wisely with fixed tasks are likely to violate safety constraints in different task settings. 
In other words, for AI to truly mirror human intelligence, it must possess the ability to adapt quickly to new tasks under constraints. Therefore, it is crucial to develop learning algorithms that enable LES to rapidly adapt while adhering to safety specifications. As it is still challenging for existing safe learning approaches, our goal here is to bridge such a gap by achieving a fast adaptation regarding both performance and safety guarantees in non-stationary environments.


Specifically, we investigate fast-adapting safe RL through the lens of meta-learning, which admits a bi-level structure. Constrained Policy Optimization (CPO, \cite{achiam2017constrained}) is employed as the base module for the updates of task-specific parameters and meta parameters at the inner and outer levels, respectively.
However, it is well-known that unconstrained meta-learning already admits complex differentiation of the loss function with respect to the meta parameter. This issue is even worse in constrained meta-learning since the inner-level updates under constraints complicate the differentiation process even further. To tackle this challenge, we use CPO which convexifies the constrained policy learning within a trust region for policy updates. Additionally, to facilitate efficient end-to-end differentiation for effective meta-training, we employed Differentiable Convex Optimization (DCO, \cite{cvxpylayers}) to bridge the inner-level and outer-level updates. As convexity enhances the efficiency of both forward pass and backpropagation, this framework will support adaptable safety guarantees at scale under non-stationarity. This will be further explored in Section \ref{Cvxpylayers}, where we delve into solving a constrained meta-learning problem for unprecedented testing tasks in non-stationary environments.
To the best of our knowledge, this is the first attempt to build such a framework, providing a promising solution for fast adaptation with safety specifications.



Our main contributions are listed as follows.
\begin{itemize}
    \item Building a novel architecture by integrating constrained RL into the meta-learning framework, enhancing the ability to provide adaptable safety guarantees.
    \item Developing a practical method to solve constrained metal-RL via successive convexification and DCO for end-to-end trainability.
    \item Conducting a thorough evaluation of the Meta-CPO algorithm, which outperforms the benchmarks regarding performance and safety satisfaction.
\end{itemize}

\section{Preliminaries}
\subsection{Constrained Reinforcement Learning}\label{sec:rl}
A Markov Decision Process (MDP) serves as a mathematical framework for modeling sequential decision-making problems. It is composed of several key components represented as a tuple: $(S, A, R, P, \mu)$. Here, $S$ denotes the set of states, $A$ represents the set of actions, $R : S \times A \times S \rightarrow \mathbb{R}$ signifies the reward function, $P : S \times A \times S \rightarrow [0, 1]$ represents the transition probability function. Specifically, $P(s' | s, a)$ indicates the probability of transitioning to state $s'$ given the action $a$ the agent took in previous state $s$. Additionally, $\mu : S \rightarrow [0,1]$ denotes the starting state distribution. Within an MDP, a policy $\pi : S \rightarrow P(A)$ refers to a mapping from states to probability distributions over actions. The notation $\pi(s | a)$ implies the probability of selecting action $a$ in state $s$. 

While RL only aims to maximize a cumulative discounted reward 
$J_{R}(\pi) = \mathbb{E}_{\tau \sim \pi} \left[ \sum_{t=0}^{\infty}\gamma^t R(s_t, a_t) \right]$, 
constrained RL will additionally enforce a cumulative discounted cost constraint as
$J_{C}(\pi) = \mathbb{E}_{\tau \sim \pi} \left[ \sum_{t=0}^{\infty} \gamma^t  C(s_t, a_t) \right] \leq h,$
where $h$ is the safety threshold, $C(s_t, a_t)$ is the cost function, $\gamma$ the discount factor, $\tau$ is the trajectory $\tau = (s_0 , a_0 , s_1 , \dots )$, and $\tau \sim \pi$ means that the trajectory distribution depends on $\pi$ in the following way: $s_0 \sim \mu, a_t \sim \pi(a_t|s_t), s_{t+1} \sim P(s_{t+1} |s_t , a_t )$.

To guide effective policy learning, one can use either an action-value function, $Q^{\pi}_{R}(s,a)$, or a state-value function, $V^{\pi}_{R}(s)$. These functions estimate the expected future return based on the current state and chosen action (action-value) or just the state itself (state-value). Their formal definitions are $Q^{\pi}_{R}(s,a) = \mathbb{E}_{\tau \sim \pi} \left[ \sum_{t=0}^{\infty} \gamma^tR(s_t,a_t)|s_0 = s, a_0 = a \right]$ for action-value function and $V^{\pi}_{R}(s) = \mathbb{E}_{\tau \sim \pi} \left[ \sum_{t=0}^{\infty} \gamma^tR(s_t,a_t)|s_0 = s \right]$ for state-value function. In analogy, the action and state value functions for the cost, $Q^{\pi}_{C}(s,a) = \mathbb{E}_{\tau \sim \pi} \left[ \sum_{t=0}^{\infty} \gamma^t C(s_t,a_t)|s_0 = s, a_0 = a \right]$ and $V^{\pi}_{C}(s) = \mathbb{E}_{\tau \sim \pi} \left[ \sum_{t=0}^{\infty} \gamma^t C(s_t,a_t)|s_0 = s \right]$, are defined similarly.


Kakade and Langford \cite{kakade} give an identity to express the performance measure of policy $\pi^{\prime}$ in terms of the
advantage function over another policy $\pi$ :
\begin{equation}
	J_R(\pi^{\prime}) - J_R(\pi) = \frac{1}{1-\gamma} \mathbb{E}_{s \sim d^{\pi^{\prime}}_{a \sim \pi^{\prime}}} \left[A_R^{\pi}(s,a)    \right]
	\label{eqn:performance}
\end{equation}
where $d^{\pi'}$ is the discounted future state distribution of policy $\pi'$, and \eqref{eqn:performance} still depends on expectation of $\pi'$. Moreover, $A^{\pi}_R(s,a)$ is the reward advantage function defined as $A^{\pi}_R(s,a) \coloneqq Q^{\pi}_R(s,a) - V^{\pi}_R(s)$. Similarly, we have the cost advantage function as $A^{\pi}_C(s,a) = Q^{\pi}_C(s,a) - V^{\pi}_C(s)$. 

Extending upon \eqref{eqn:performance}, subsequent research \cite{pmlr-v37-schulman15} defined a surrogate function, $M(\pi)$, of \eqref{eqn:performance} to bound the policy improvement, replacing dependencies on $\pi'$ to $\pi$:

\begin{equation}
    \begin{split}
        J_R(\pi^{\prime}) - J_R(\pi) \geq M(\pi^{\prime}) - M(\pi) \qquad \qquad \qquad \\
        \qquad \qquad \qquad = L_{\pi}(\pi^{\prime}) - CD^{\textrm{max}}_{\textit{KL}}(\pi,\pi^{\prime})
    \end{split}
    \label{eqn:surrogate}
\end{equation}
where $L_{\pi}(\pi^{\prime})=\frac{1}{1-\gamma} \mathbb{E}_{s \sim d^{\pi}_{a \sim \pi}} \left[A_R^{\pi}(s,a) \right]$ and $CD^{\textrm{max}}_{\textit{KL}}(\pi,\pi^{\prime})$ represents the error bound. It's worth noting that by decreasing the step size, $D^{\textrm{max}}_{\textit{KL}}$, we can minimize the error on a smaller scale, ensuring trust region updates that enforce the right-hand side (RHS) to always be positive. Constrained Policy Optimization (CPO, \cite{achiam2017constrained}) further extends this framework by introducing extra inequality constraints. This ensures the preservation of the guarantee of monotonic improvement while accommodating the additional constraint. Consequently, the resultant policies not only aim for optimal performance but also align with specified safety requirements. Building upon \eqref{eqn:surrogate}, we present a theoretical analysis that extends this methodology to a meta-learning setting.



\subsection{Constrained Policy Optimization (CPO)}\label{CPO}
\begin{figure}[t]
    \centering
    \includegraphics[width=0.35\textwidth]{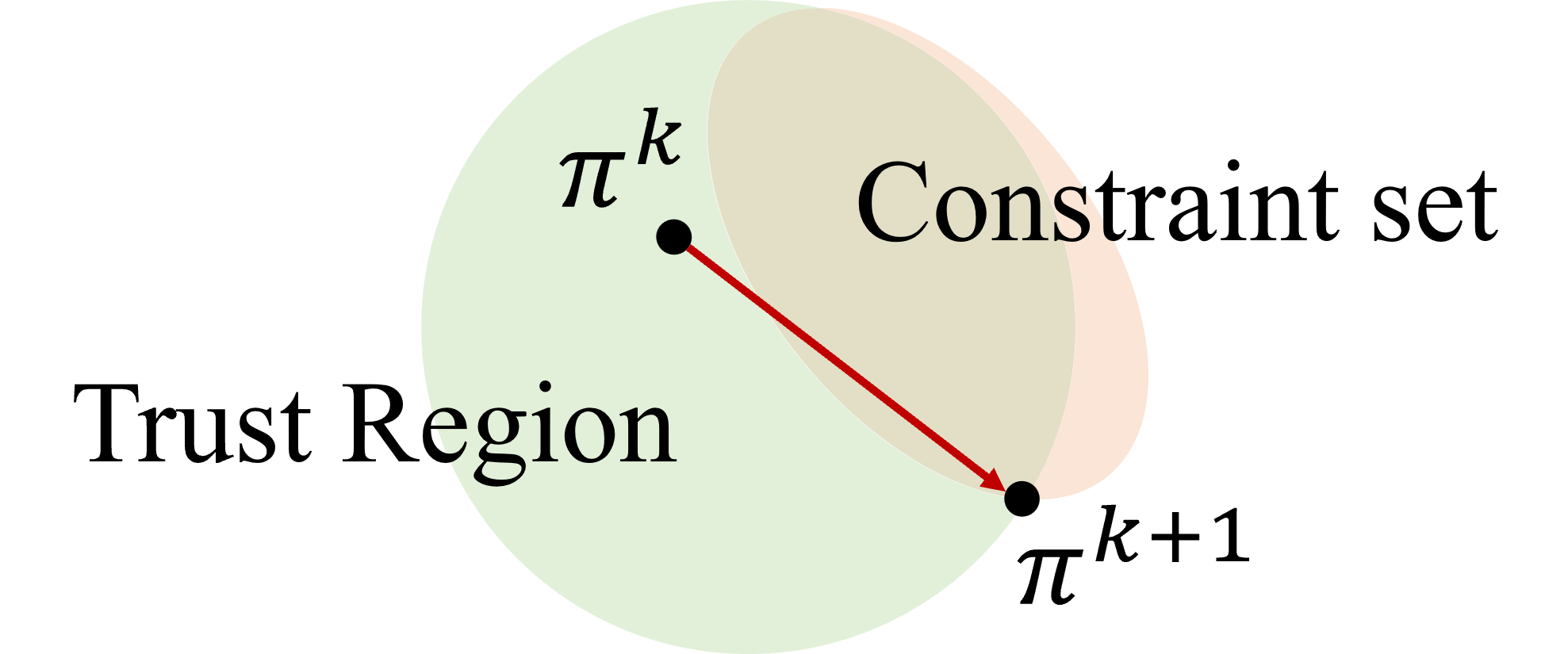}
    \caption{Update procedures for CPO \cite{achiam2017constrained}. CPO computes the update by simultaneously considering the trust region (light green) and the constraint set (light orange). 
 Figure adopted from \cite{PCPO}}
    \label{fig:CPO}
\end{figure}

Constrained RL methods (e.g., CPO) aim to balance the trade-off between achieving goals and ensuring safety in critical RL tasks, like industrial robot operation and autonomous driving. 
Leveraging the trust region for guaranteed monotonic improvement, CPO offers intuitive analytical solutions that excel at optimizing policies for a balance of rewards and safety constraints by solving a constrained optimization problem. For CPO, its update procedure is depicted in Figure \ref{fig:CPO} and the detailed update rule is formulated as follows

\begin{equation}
    \begin{aligned}    
        \theta^{k+1} = \argmax_{\theta} \quad & g^T_{\theta}(\theta - \theta^k)\\
        \textrm{s.t.} \quad & \frac{1}{2} (\theta - \theta^k)^TH(\theta - \theta^k) \leq \delta\\ 
        & b_i + a_{\theta}^T(\theta - \theta^{k}) \leq 0
    \end{aligned}
    \label{eqn:CPO1}
\end{equation}
with definitions: $g_{\theta} = \nabla_{\theta} \mathbb{E}_{s \sim d^{\pi}}\left[A^{\pi}_R(s,a)\right]$ is the gradient of the reward advantage function,	$a_{\theta} = \nabla_{\theta} \mathbb{E}_{s \sim d^{\pi}}\left[A^{\pi}_C(s,a)\right]$ is the gradient of the cost advantage function, $H$ is Hessian matrix of KL-divergence $\frac{\partial^2\bar{D}_{KL}}{\partial\theta^2}$, and  $b_i = J_i^{C}(\pi) - h$.

In case \eqref{eqn:CPO1} is infeasible, CPO instead solves
\begin{equation}
    \begin{aligned}
        \theta^{k+1} = \argmin_{\theta} \quad & a^T_{\theta}(\theta - \theta^k)\\
        \textrm{s.t.} \quad & \frac{1}{2} (\theta - \theta^k)^TH(\theta - \theta^k) \leq \delta\\ 
    \end{aligned}
    \label{eqn:CPO2}
\end{equation}
by solely decreasing the constraint value within the trust region. We will also use the following notation $\|\theta - \theta^k\|_H^2 \coloneqq (\theta - \theta^k)^TH(\theta - \theta^k) $ for simplicity.
 
In our approach, CPO serves as the primary update rule for the agents, both meta- and local ones. Because the objective of CPO is to optimize the policy while ensuring safety satisfaction, it is a good fit to be the base learner towards constrained meta-policy learning under non-stationary.


\subsection{Differentiable Convex Optimization}\label{sec:dco}
Differentiable Convex Optimization (DCO,~\cite{cvxpylayers}) is a framework that aims to facilitate the differentiation of convex optimization problems. The DCO layers provide a framework for expressing and solving convex optimization problems in a way that allows for efficient gradient computations and integration with deep learning models. These layers offer a differentiable representation of convex optimization problems, enabling the computation of gradients with respect to the optimization variables. By utilizing affine-solver-affine (ASA) composition and employing canonicalization techniques, DCO layers ensure straightforward computation of gradients through the backward pass. This advancement allows for the integration of convex optimization with deep learning models, enabling meta-learning with nested convex optimization modules with learnable parameters. 
Specifically, the ASA consists of taking the optimization problem's objective and constraints and mapping them to a cone program.
For the following general quadratic programming (QP)

\begin{equation}
\begin{aligned}
\min_x &\ \  \frac{1}{2} x^T Q x+q^T x 
\text { s.t.} \ \  A x=b , G x \leq h,
\end{aligned}
\end{equation}
we can write the Lagrangian function of the problem as:
\begin{equation}
L(z, \nu, \lambda)=\frac{1}{2} z^T Q z+q^T z+\nu^T(A z-b)+\lambda^T(G z-h)
\end{equation}
where $\nu$ are the dual variables on the equality constraints and $\lambda \geq 0$ are the dual variables on the inequality constraint.
Using the KKT conditions for stationarity, primal feasibility, and complementary slackness.
\begin{equation}
\begin{aligned}
Q z^{\star}+q+A^T \nu^{\star}+G^T \lambda^{\star} & =0 \\
A z^{\star}-b & =0 \\
D\left(\lambda^{\star}\right)\left(G z^{\star}-h\right) & =0
\end{aligned}
\end{equation}
By differentiating these conditions, we can shape the Jacobian of the problem as follows.
\begin{equation}\nonumber
\left[\begin{array}{l}
d_z \\
d_\lambda \\
d_\nu
\end{array}\right]=\left[\begin{array}{ccc}
Q & G^T D\left(\lambda^{\star}\right) & A^T \\
G & D\left(G z^{\star}-h\right) & 0 \\
A & 0 & 0
\end{array}\right]^{-1}\left[\begin{array}{c}
\left(\frac{\partial \ell}{\partial z^{\star}}\right)^T \\
0 \\
0
\end{array}\right]
\end{equation}
Furthermore, via chain rule, we can get the derivatives of any loss function of interest regarding any of the parameters in the QP.

\section{Constrained \\ Meta-Reinforcement Learning} \label{metaCPO}
To leverage previous learning experiences, it is crucial to train a model that can adapt to multiple tasks rather than only being optimized for a single task. This is where meta-learning comes into play. Model-agnostic meta-learning (MAML) \cite{MAML} has emerged as a powerful technique for training models with generalization capabilities over unseen tasks, demonstrating its potential for few-shot adaptation in both supervised learning and reinforcement learning domains.

 In MAML, there are two key components: the meta-learner ($\theta$) and the local-learner ($\phi$). Meta-learner could be a distinct model parameter or could be other adaptable parameters such as learning rate and, in reinforcement learning, discount factor $\gamma$ \cite{Meta-learning:survey}. The goal of meta-learning is to train the meta-learner in a way that it can quickly adapt to new tasks with minimal training, while the local learner represents the updated parameters obtained through one or more updates on a specific task. By averaging the updates of the local-learner across multiple tasks, the model can achieve strong generalization over its given training tasks. 

Each task $\mathcal{T}_i$ consists of its own initial state distribution, $\mu_i(s_0)$ and loss function $\mathcal{L}_{\mathcal{T}_i}$, which leads to the task-specific advantage function $A^{\pi}_{i,R}$. Just like RL introduced in \Cref{sec:rl}, $\mathcal{T}_i$ is represented as an MDP with horizon $H$, and its trajectory rollout is used for policy evaluation and updates. With the reward and cost functions associated with $\mathcal{T}_i$ and a parameterized policy $\pi_\theta$, the corresponding return and cost gradients can be expressed similarly to the CPO formulation:
$g(\theta, \mathcal{T}_i) = \mathbb{E}_{s_t, a_t \sim \pi_{\theta}, \mu_{\mathcal{T}_i} } \left[\nabla_\theta A^{\pi}_{i,R}(s, a|\theta)\right]$
and
$a(\theta, \mathcal{T}_i) =  \mathbb{E}_{s_t, a_t \sim \pi_{\theta}, \mu_{\mathcal{T}_i} } \left[\nabla_\theta A^{\pi}_{i,C}(s, a|\theta)\right]$
.



\begin{figure}[t]
    \centering
    \includegraphics[width=0.475\textwidth]{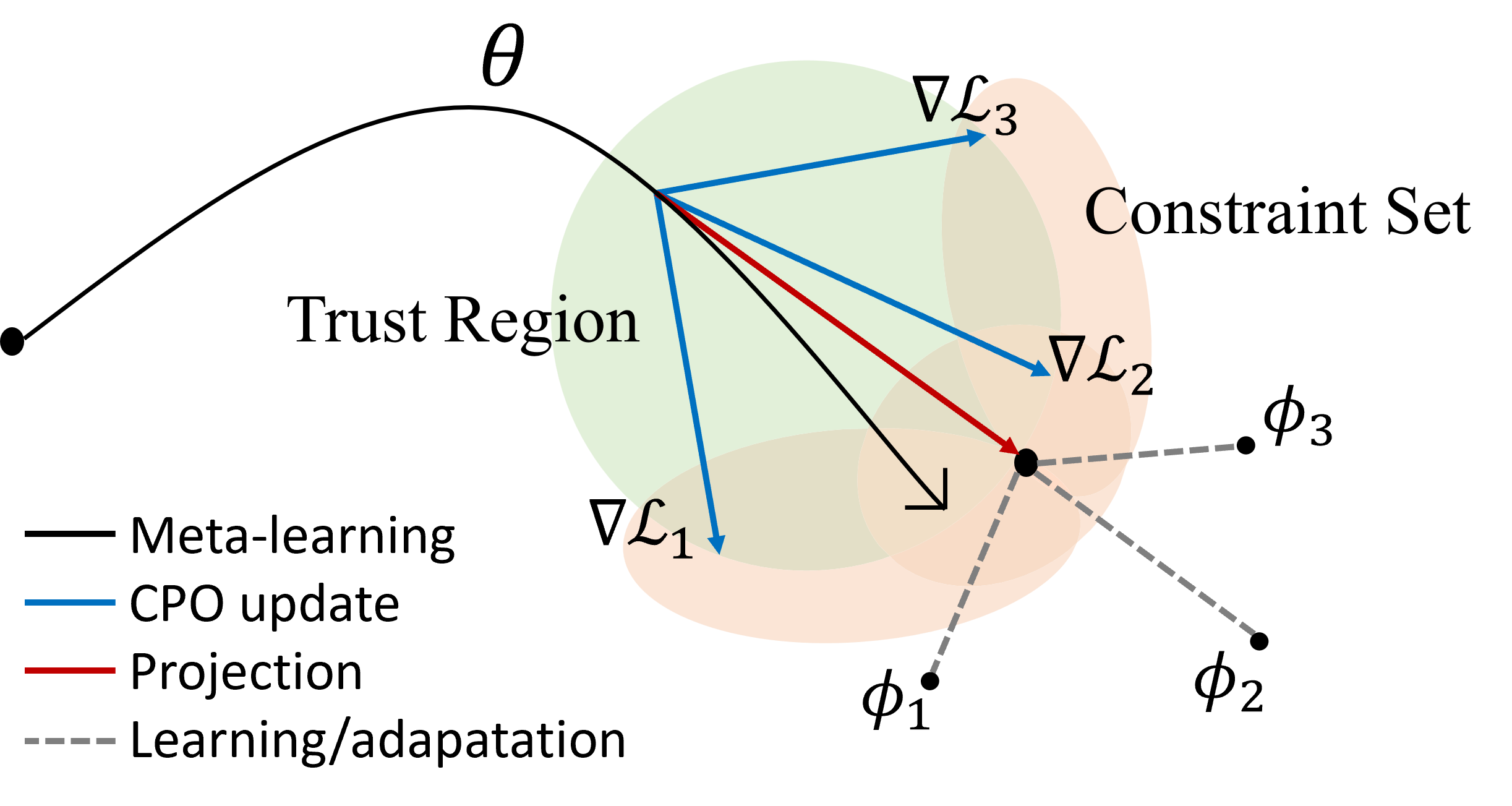}
    \caption{The black curve under meta parameter $\theta$ is the learning trajectory in meta-policy space. Each blue $\nabla L_i=\nabla_\theta A_{\phi_k}^R$ is the task-specific gradient, performing a tug-of-war of each task to fulfill generalization/ adaptation over multiple tasks. The projection finds a better intersection of the constraint set, preventing safety violations arising from average gradients.}
    \label{fig:maml}
\end{figure}

In general, meta-learning trains meta-parameter $\theta$ in the outer level and task-specific parameters $\phi_i$ in the inner level for task $\mT_i$. Meta parameters can produce task-specific parameters (e.g., as an initializer) for fast adaptation for both training and testing in the same fashion, specified as follows
\begin{equation}\label{eq:meta}
\begin{aligned}
    &\texttt{outer-level:}  \\
    &\qquad \theta \coloneqq \argmax_{\theta} F(\theta), \quad \textrm{s.t.} \quad G(\theta) \le 0 \\
    &\qquad\qquad \text{where}\ \  F(\theta) = \frac{1}{M}\sum_{i=1}^M A^{\pi}_{i,R}(\phi_i, \mD_i^{tr}), \\
    &\texttt{inner-level:} \\
    &\qquad \phi_i = \text{Alg}(\theta, \mD_i^{tr}) = \textbf{CPO}(\theta, \mD_i^{tr}),
\end{aligned}
\end{equation}
where $G(\theta)$ is defined in the same way as $F(\theta)$ with the defined $A^{\pi}_{i,C}(\phi_i, \mD_i^{tr})$. 
Moreover, $\mD^{tr}$ and $\mD^{test}$ are the collection of training and testing tasks respectively. In the unconstrained settings, the $\text{Alg}(\bullet)$ is often instantiated as gradient descent updates, which is however inapplicable in constrained settings.
Moreover, in the inner level, $\text{Alg}(\theta, \mD_i^{tr})$ can be executed multiple times. 
Model-based or model-free, primal methods or primal-dual methods, provide many options for the instantiation of $\text{Alg}(\bullet)$. Here, constrained policy optimization (CPO,~\cite{achiam2017constrained}), a model-free primal method, is chosen as the optimizer. 

Our algorithm can be broken down into two parts: local updates with nominal CPO steps and meta updates following the differentiation through the meta parameters. As previously mentioned, CPO serves as a primary update rule for both parameters. 
This sequential process ensures that the policy optimization is enhanced through the local learner's update and generalized effectively by the meta-learner, resulting in improved overall performance and adherence to the specified constraints. We elaborate on our approach in the subsequent section, and Figure \ref{fig:maml} provides a visual representation of the MAML concept in \textit{constrained} settings. In this context, multiple local learners engage in a tug-of-war to guide the policy to their respective task's optimal point. The projection then identifies the intersection of constraint sets from each task, helping to mitigate safety violations that average updates might cause.

\begin{figure}[t]
    \centering
    \includegraphics[width=0.475\textwidth]{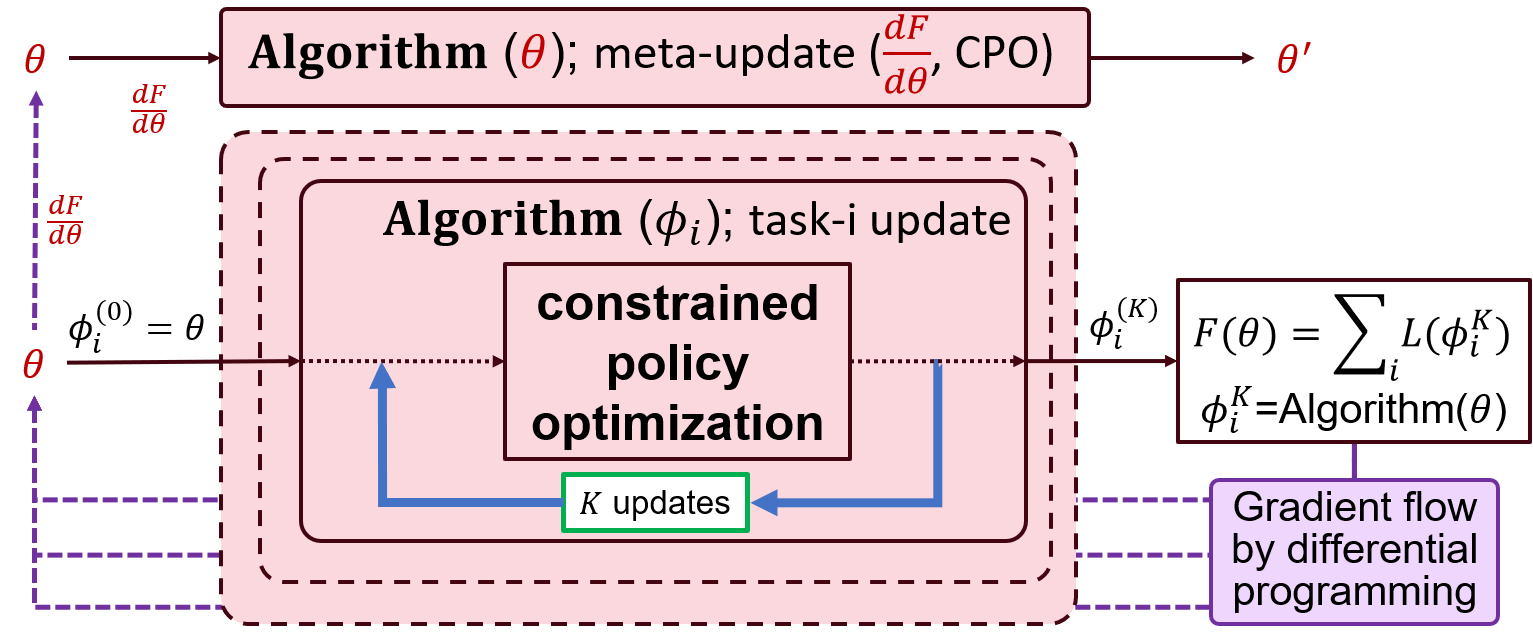}
    \caption{Diagram of our meta-learning approach with CPO as the base algorithm. It optimizes for meta-parameter $\theta$ that can quickly adapt to new tasks \textit{under constraints}.}
    \label{fig:metamaml}
\end{figure}




\section{End-to-End Trainability via DCO \\ with Adaptable Safety Guarantee} \label{Cvxpylayers}

In meta-learning, the main computational complexity comes from computing gradients of the loss function with regard to the meta parameter, which has to go through local updates. However, such gradient computations may be intractable due to the complex optimization problem, \emph{i.e.} CPO. To enable meta-learning and facilitate differentiation within the optimization, the Differentiable Convex Optimization (DCO) is used. This framework allows us to effectively and efficiently enable end-to-end differentiation within the optimization layers. 

\subsection{Local Update (Inner-level)}
When adapting to a certain task $\mathcal{T}_i$, we update the model’s meta parameters $\theta$ to its local copy $\phi_i$ with CPO. In our method, the updated parameter vector $\phi_i$ is computed using multiple CPO updates on task $\mathcal{T}_i$. With $\phi_i^0 = \theta$, the local parameter $\phi_i$ is updated successively in the following form
\begin{equation}
    \begin{aligned}
        \phi_i^{k+1} = \argmax_{\phi_i} \quad & g(\phi_i^k, \mD_i^{tr})^T(\phi_i - \phi_i^k)\\
        \textrm{s.t.} \quad & \frac{1}{2} \|\phi_i - \phi_i^k\|_H^2 \leq \delta\\ 
        & b_{\phi_i} + a(\phi_i^k, \mD_i^{tr})^T(\phi_i - \phi_i^{k}) \leq 0
    \end{aligned}
    \label{eqn:local_CPO}
\end{equation}
where the superscript of $\phi_i^{k}$ represents the iteration index of local updates.

The parameters are defined in a similar way to those in \eqref{eqn:CPO1}. 
To obtain the local-learner $\phi_i^{K}$ after $K$ updates, we implement CPO for each task $\mathcal{T}_i$ to perform local updates. During the local-update, individual gradients and final updates are stored to link each update at the $K$ local step to the meta-learner using respective gradients that find average updates capturing shared knowledge across tasks. However, policy updates may violate cost constraints $J_{C}(\pi) \leq h$ and KL-divergence $|\pi - \pi^k|_H^2$, requiring a \textit{backtracking line search} for feasibility~\cite{achiam2017constrained}."

\subsection{Meta Update (Outer-level)}

Recall the meta-learning framework in \eqref{eq:meta} with the following meta loss function with regard to meta parameter $\theta$
\begin{equation}\label{eq:metaloss}
\begin{aligned}
    \max_{\theta} \quad & F(\theta) = \frac{1}{M}\sum_{i=1}^M A^{\pi}_{i,R}(\phi_i, \mD_i^{tr}), \\
    \textrm{s.t.} \quad & G(\theta) \le 0
\end{aligned}
\end{equation}

\begin{algorithm}[t]
	\caption{\textbf{Meta-CPO} for fast adaption under constraints}\label{alg:cap}
	\begin{algorithmic}[1]
		\REQUIRE $p(\mathcal{T})$: distribution over tasks
		\REQUIRE $\delta, h$: optimization constraints
		\STATE \textbf{Initialization} Randomly initialize $\theta$
		\WHILE{not done}
        \STATE \texttt{/*local updates*/}
        \STATE Sample tasks under $p(\mathcal{T})$
		\FOR{each sampled $\mathcal{T}_i$}
            \FOR{$k=0,\ldots,K-1$}
    		\STATE Sample multiple trajectories $\mathcal{D} = \{\tau\}$ using policy $\pi_{\phi_i^k}$ in $\mathcal{T}_i$ with $\phi_i^0=\theta$
                \STATE Estimate parameters in \eqref{eqn:local_CPO}
    		\STATE Update $\phi_i^{k+1} \leftarrow \phi_i^{k}$ with CPO in \eqref{eqn:local_CPO} and backtracking linesearch
            \ENDFOR
		\ENDFOR

        \STATE \texttt{/*meta updates*/}
  	\STATE Estimate $F, G$ and compute the gradients $d F/d\theta$ and $d G/d\theta$ with \eqref{eq:dF}      
               using DCO
		\STATE Update $\theta$ with \eqref{eqn:meta_CPO} and backtracking line search

		\ENDWHILE
        \STATE \textbf{Output:} $\theta$ for meta-testing (\textit{not shown}) in new tasks.
        
	\end{algorithmic}
\end{algorithm}

To update the meta parameter $\theta$ by gradient descent algorithms, the gradient can be estimated by (chain rule) $\frac{d F}{d\theta}  = \frac{1}{M}\sum_{i=1}^M {\frac{d \text{Alg}_i(\theta)}{d\theta}} g(\phi_i, \mD_i^{tr})$. Note that the total derivative $\frac{d \text{Alg}_i(\theta)}{d\theta}$ passes derivatives through $\text{Alg}_i(\bullet)$ such that we need to differentiate through it. This highlights a major challenge in meta-learning: even when $\text{Alg}(\bullet)$ is in a fairly simple form of gradient descent in \eqref{eq:meta}, the requirement for second-order derivative makes it computationally intense. As a result, \cite{MAML} only keeps the first-order terms, and \cite{rajeswaran2019meta} proposes implicit differentiation.
However, for safe learning, the safety constraint can not be reconciled, such as a soft constraint/penalty instead of a hard one. In this case, the complex issue of differentiation is made worse by the fact that $\text{Alg}(\bullet)$ solves a constrained learning problem. This can be part of the reason why constrained meta-learning has not been investigated as much as its unconstrained counterpart because the constraints might damage or complicate end-to-end differentiability. Here we propose to differentiate through the constrained policy update, via DCO, to enable end-to-end meta-training. 

With DCO in \Cref{sec:dco}, we can obtain the derivative of the meta loss function with regard to the meta parameters as
\begin{equation}\label{eq:dF}
\frac{d F}{d\theta}  = \frac{1}{M}\sum_{i=1}^M {\prod_{k=0}^{K-1} \frac{d \text{Alg}_i(\phi_i^{(k+1)})}{d\phi_i^{(k)}}}g(\phi_i^K, \mD_i^{tr})
\end{equation}
where $\frac{d \text{Alg}_i(\phi_i^{k+1})}{d\phi_i^{k}}$ is enabled by differentiating through the local CPO update in \eqref{eqn:local_CPO} by DCO, which allows computing the derivative of the loss function with respect to any parameters in the quadratic programming \eqref{eqn:local_CPO}. As multiple parameters in \eqref{eqn:local_CPO}, appearing in both objective and constraints, depend on $\phi_i^{k}$, the total derivative ${d\text{Alg}_i(\phi_i^{k+1})}/{d\phi_i^{k}}$ will be the summation of all of the partial derivatives. In analogy, $\frac{d G}{d\theta}$ can be computed in the same way for the update in \eqref{eq:metaloss}.

With the derivative of the meta objective and constraint functions evaluated, the meta updates can be readily performed in a similar way using CPO. 

\begin{equation}
    \begin{aligned}
        \theta^\prime = \argmax_{\theta^\prime} \quad & \bigg(\frac{d F}{d\theta}\bigg)^T (\theta^\prime - \theta)\\
        \textrm{s.t.} \quad & \frac{1}{2} \|\theta^\prime - \theta\|_H^2 \leq \delta_\theta\\ 
        & b_\theta + \bigg(\frac{d G}{d\theta}\bigg)^T(\theta^\prime - \theta) \leq 0
    \end{aligned}
    \label{eqn:meta_CPO}
\end{equation}

Once \eqref{eqn:meta_CPO} is infeasible, a similar strategy to \eqref{eqn:CPO2} is adopted. The evaluation of $F(\theta)$, $G(\theta)$, and their derivatives over multiple tasks enables the generalization to new tasks, including both return improvement and the satisfaction of the safety constraints. 
The whole architecture of meta-learning with CPO (Meta-CPO) is depicted in \Cref{fig:metamaml}. The pseudo-code is provided in Algorithm \ref{alg:cap}, while the complete source code is available on GitHub\footnote{https://github.com/Mgineer117/Meta-CPO}.

\subsection{Theoretical Analysis}
In our approach, the rollout $[\theta_n \cdots \bigl\{ \phi^k_i \bigr\}_{i=1}^M \cdots \theta_{n+1}]$ is made to optimize meta-learner $\theta_n$ with differentiation through local-learners $\phi^k_i$, where $k$ is the number of local iterations and $i$ is index of local-learners. For bi-level meta-updates, we reformulate the work of TRPO/CPO
for the theoretical analysis of meta-learner $\theta$. We begin with defining the average performance of local-learner: 
$\Delta \Bar{J}^{k+1}:=\Bar{J}(\phi^{k+1})-\bar{J}(\phi^{k})
    \geq \frac{1}{M}\sum_{i=1}^M \bigl[L_{\phi_i^{k+1}}(\phi_{i}^k) - C_i^kD^{\textrm{max}}_{\textit{KL}}(\phi_i^k,\phi_{i}^{k+1})\bigr]$,
where $\bar{J}(\phi^{k+1})$ is a mean performance of $\phi_{i}$ at local step $k+1$ (i.e., $\frac{1}{M} \sum_{i=1}^MJ(\phi^{k+1}_{i})$), and RHS is a mean surrogate of performance difference of $\phi^k_i$ and its next update $\phi^{k+1}_{i}$. 
Since all local learners demonstrate improvement over their previous iterations within the trust region, carefully chosen step sizes ensure that the meta-learner's performance $J(\theta_{n+1})$ is non-decreasing compared to the previous meta-learner's performance $J(\theta_{n})$. This property implies that as long as the local updates stay within the meta-learner's trust region, the meta-learner update is guaranteed to be superior or at least the same as a local learner, $J(\theta_{n+1}) \geq \Bar{J}(\phi^{k+1}_{i})$. Hence, this can yield the performance guarantees of meta parameters $\theta$:
$$J(\theta_{n+1}) - J(\theta_n) \geq \Delta \Bar{J}^{k+1} \geq \Bar{L}_{\phi_i^k}(\phi_{i}^{k+1}) - \Bar{C}_i^k\Bar{D}^{\textrm{max}}_{\textit{KL}}$$
All above analysis upon the return maximization, $J(\theta)$, applies for cost satisfaction, $J_C(\theta)$, as CPO does. This also aligns with empirical experiments of Meta-CPO with a steady learning curve.
Consequently, our approach can achieve adaptable safety guarantees while maintaining monotonic performance improvement.

\subsection{Limitations}
However, current DCO layers (cvxpylayers\footnote{https://github.com/cvxgrp/cvxpylayers}) have limited ability to handle a large number of parameters in computations. This restricts our approach to a smaller parameter scale resulting in unstable performance for high-dimensional tasks and external disturbances. Additionally, the use of cvxpylayers prevents us from employing certain mathematical tricks for effective and memory-efficient matrix-vector product computations for solving CPO. Accordingly, we changed the KL-divergence metric to the Euclidean metric as $ \|\theta - \theta^k\|_H^2 \longrightarrow ||\theta - \theta^{k}||^2_2$. Understanding that such conversion results in model-variant updates can compromise the guaranteed monotonic improvement within the trust region, we implement multiple local steps to alleviate this issue. This approach ensures that the local learner propagated from the meta-learner, is a better policy (under the mild assumption), aligning with the theoretical framework we propose.







\section{Experiment} \label{experiments}
\begin{figure}[t]
    \centering
    \begin{minipage}{0.45\columnwidth}
        \centering
        \includegraphics[width=0.95\columnwidth]{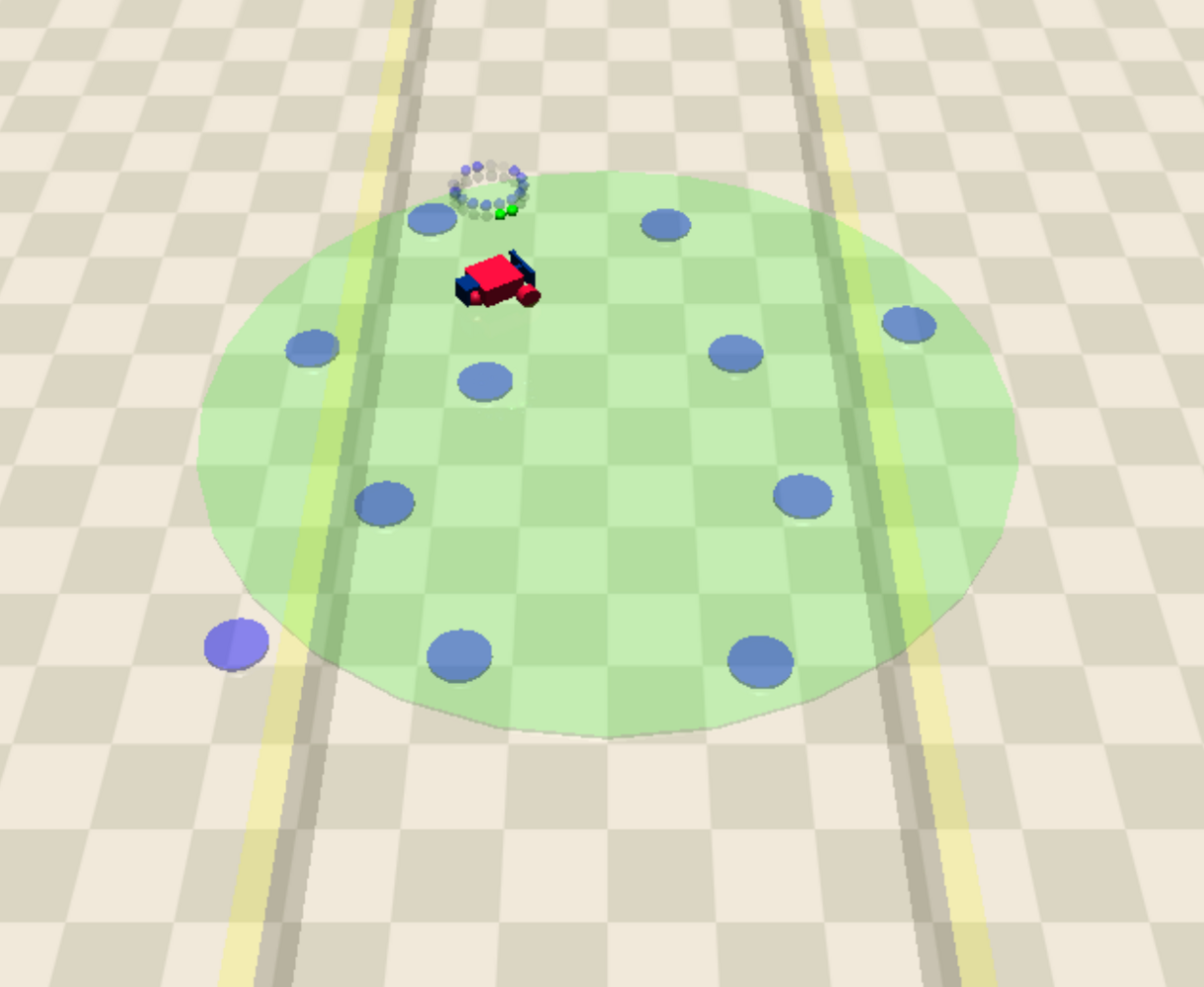}
        \caption*{(a) Car-Circle-Hazard}
        \label{fig:figure1}
    \end{minipage}%
    \begin{minipage}{0.45\columnwidth}
        \centering
        \includegraphics[width=0.95\columnwidth]{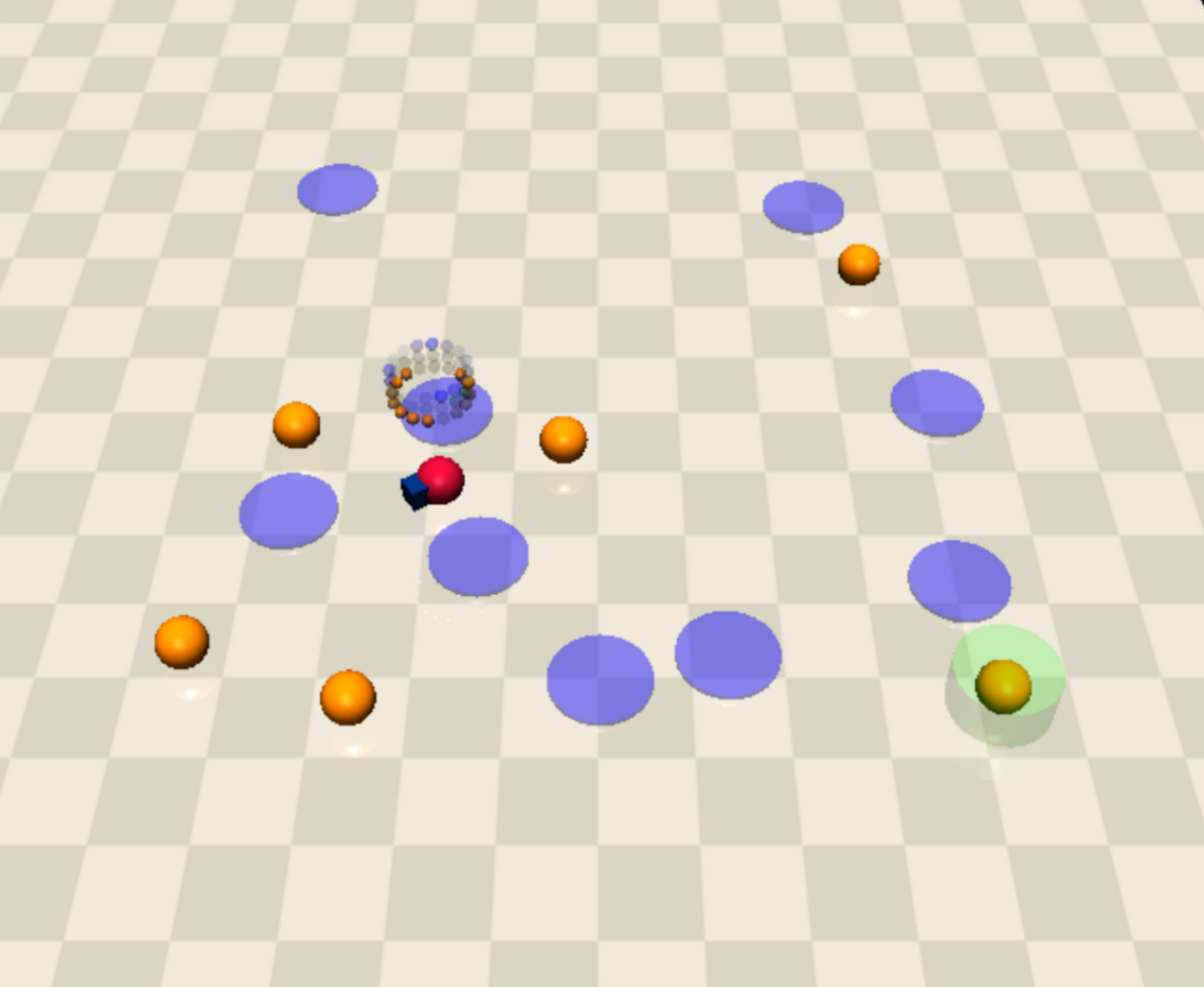}
        \caption*{(b) Point-Button}
        \label{fig:figure2}
    \end{minipage}
    
    \caption{Car-Circle-Hazard and Point-button environments. In Car-circle-Hazard environment, the agent avoids the walls and hazards, while the agent strives to only touch activated button by avoiding other dead buttons and hazards in Point-Button environment.}
    \label{fig:envs}
\end{figure}

\begin{figure*}[t]
    \centering
    \begin{tabular}{c c c}
        \multicolumn{3}{c}{\includegraphics[width=0.45\textwidth]{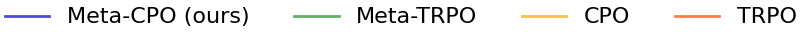}}\\
        \includegraphics[width=0.315\textwidth]{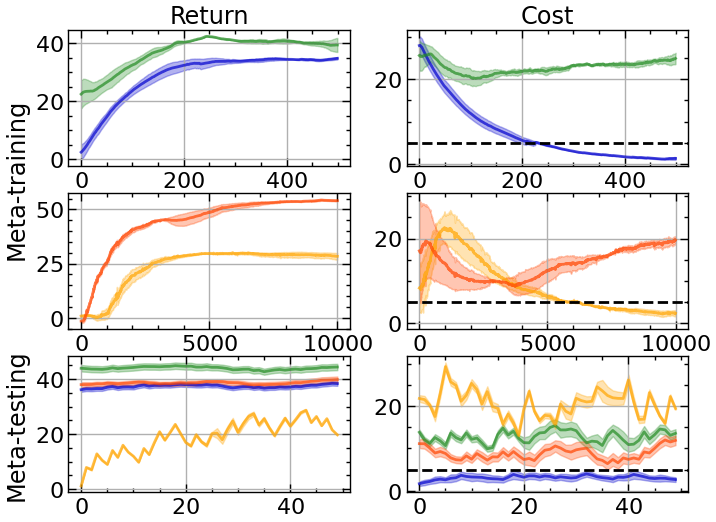}    &
        \includegraphics[width=0.315\textwidth]{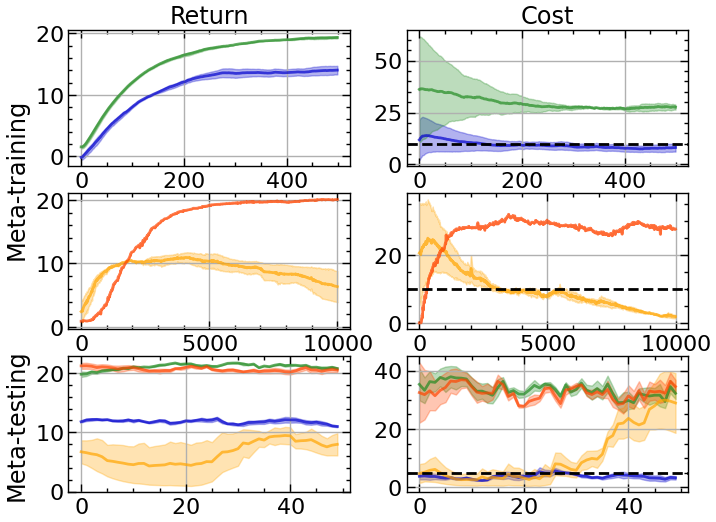}    &
        \includegraphics[width=0.315\textwidth]{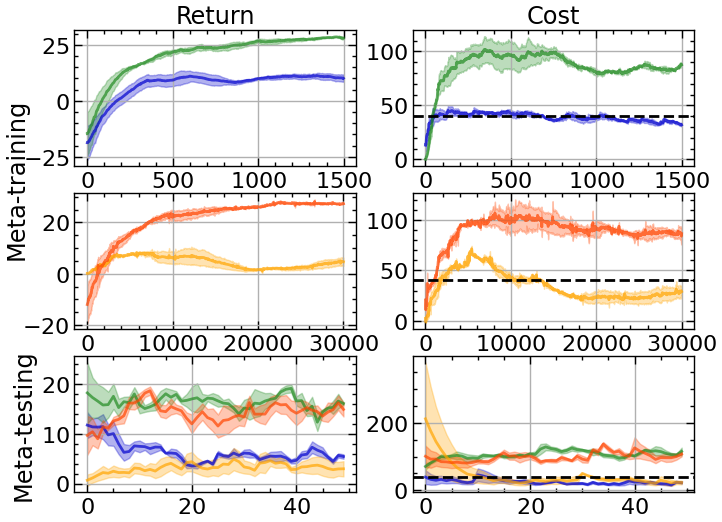}    \\
        (a) Point-Circle & (b) Car-Circle-Hazard & (c) Point-Button \\
    \end{tabular}
    
    \caption{Two columns for each task: return (higher is better)  and cost (lower is better). The black dashed line is the cost upper bound (below the line means satisfaction).
    The first two rows are training for meta and non-meta algorithms, respectively.
    The third row is meta-testing for \textit{unseen} tasks, i.e., deploying meta-trained policies with few-shot adaptation and evaluating them in return and cost. All methods are trained with 3 random seeds, and the mean (solid curve) and standard error (error bar) are plotted with training iterations on $x$-axis. \textbf{Meta-CPO (ours)} works well in testing against baselines. Specifically for Car-Circle-Hazard, the cost upper bound was changed tighter in testing, i.e., $h: 10 \rightarrow 5$; see Figure \ref{fig:whole_data} (b) (second column, black dashed line). Meta-CPO (ours) can adapt to satisfy the safety constraints (below limit) while all others behave unstably or fail.}
    \label{fig:whole_data}
\end{figure*}

In our experiments, we aim to answer the following:
\begin{itemize}
	\item Does Meta-CPO successfully achieve safety satisfaction in a fast-adapting manner?
	\item How much is Meta-CPO improving the test results?
	\item What major benefits are gained by Meta-CPO?
\end{itemize}

We designed three environments utilizing the Python Safety Gym library \cite{, SafetyGym1, SafetyGym2}.
Presented below are concise descriptions of tasks within the environments:

\begin{itemize}
	\item \textbf{Point-Circle:} The agent is rewarded for running in a circle, but is constrained to stay within a safe region.
        \item \textbf{Car-Circle-Hazard:} The agent is rewarded for running in a circle, while staying within a safe region and avoiding hazards.
	\item \textbf{Point-Button:} The agent is rewarded for touching a goal button, but is constrained to touch any no-goal button and step on hazards.
\end{itemize}

The specifics of each environment, featuring non-physical Walls and Hazards, are visually illustrated in Figure \ref{fig:envs}. Our experimentation includes three distinct environmental configurations: point-circle ($S \subseteq \mathbb{R}^{28}, A \subseteq \mathbb{R}^{2}$), car-circle ($S \subseteq \mathbb{R}^{56}, A \subseteq \mathbb{R}^{2}$), and point-button ($S \subseteq \mathbb{R}^{60}, A \subseteq \mathbb{R}^{2}$). We used a policy network with two hidden layers of (32, 16). Larger networks like (64, 32) are also feasible following experimental settings demonstrated by the authors of CPO, albeit with some computational cost trade-off.

At each iteration of meta-learning, five tasks are sampled and five local updates are performed for each task (i.e., $K=5$).
Each task within these environments was generated with unique parameters, including factors like radius, distance between walls, number of hazards, and the range for spawning objects. These parameters were selected randomly from uniform distributions within predefined ranges. Table \ref{tab:table1} provides a comprehensive overview of the specifics.
Following meta-training, a meta-testing phase was conducted to evaluate the rapid learning capabilities of the meta-learner with unseen tasks.

\subsection{Meta-CPO Evaluation and Comparative Analysis}
The learning curves depicting the progress of meta-training and meta-testing are presented in Figure \ref{fig:whole_data}. To establish a benchmark, we have included Meta-CPO, Meta-TRPO, CPO\footnote{https://github.com/SapanaChaudhary/PyTorch-CPO}, and TRPO in our comprehensive analysis. 

Our analysis indicates that the Trust Region Policy Optimization (TRPO) method consistently converges towards high returns but exhibits significant constraint violations, as expected. Conversely, CPO demonstrates notably unstable behavior during testing phases. The meta-algorithm consistently outperforms non-meta algorithms in both training and testing phases, showcasing rapid and robust adaptation to distinct tasks.

Our innovative Meta-CPO algorithm excels in rapidly learning new tasks while simultaneously satisfying safety constraints. Not only can it acquire new skills swiftly while ensuring safety, but it also demonstrates remarkable adaptability to varied cost constraints. As illustrated in Figure \ref{fig:whole_data} (b), an agent trained with a cost limit of $h = 10$ seamlessly transfers its knowledge to operate effectively under a tighter limit of $h = 5$. This makes Meta-CPO a robust choice in non-stationary environments where safety is paramount.

 \begin{table}[!t]
 \centering
    \begin{tabular}{c|c|c}
    \hline
    Envs.& \multicolumn{2}{c}{Meta-training} \\ \hline 
     P-Circle & $1.0 \leq r_c \leq 1.5$ & $0.65 \leq s \leq 0.75$ \\ \hline
    C-Circle-H & $0.7 \leq r_c \leq 1.0$ & $3 \leq n_h \leq 7$ \\ \hline
    P-Button & $3 \leq (n_b, n_h) \leq 6$ & $1.75 \leq r_s \leq 2.0$  \\ \hline \hline

     Envs. & \multicolumn{2}{c}{Meta-testing} \\ \hline 
     P-Circle & $2.0 \leq r_c \leq 2.5$ & $0.55 \leq s \leq 0.65$ \\ \hline
    C-Circle-H &  $1.2 \leq r_c \leq 1.5$ & $7 \leq n_h \leq 12$ \\ \hline
    P-Button &  $6 \leq (n_b, n_h) \leq 10$ & $2.5 \leq r_s \leq 3.0$ \\ \hline

    \end{tabular}
    \caption{P, C, and H represent Point, Car, and Hazard, respectively. Tasks are specified with different environmental and safety settings with the following parameters: $r_c$ and $r_s$ denote the circle radius and spawning range of objects, while $s$ is the wall distance scale. Additionally, we have $n_b$ for the number of buttons and $n_h$ for the number of hazards. For P-Circle and C-Circle-H, $s \times r_c$ determines the wall distance, and the same $s$ is applied for both environments. Task-specific parameters were sampled under a uniform distribution and no testing tasks were seen during training.}
    \label{tab:table1}
 \end{table}

\section{Related Works} 
There is a large volume of works on safe/robust learning, including ~\cite{zhang2020robust1, zhang2020robust}, robotic learning~\cite{brunke2021safe, singh2021reinforcement}, and comprehensive surveys~\cite{garcia2015comprehensive, moos2022robust}. Specifically, the uncertainty variable can be treated as a context variable representing different tasks and can be subsequently solved as multi-task or meta-learning problems~\cite{eghbal2021learning, rakelly2019efficient}. Moreover, given optimization theories, robust learning algorithms have also been developed based on interior point method~\cite{jin2021safe, liu2020ipo}, successive convexification~\cite{achiam2017constrained} and (augmented) Lagrangian methods~\cite{bertsekas2015parallel, geibel2005risk, chow2017risk, chow2018lyapunov, stooke2020responsive}. In learning-based control, Lyapunov theory, model predictive control, and control barrier functions are also employed to develop robust learning algorithms~\cite{choi2020reinforcement, zheng2021safe, cheng2019end, ames2016control, berkenkamp2017safe, sun2021fisar, chriat2023optimality, chriat2023wasserstein, chriat2023distributionally, kanellopoulos2021temporal}. Additionally, with the worst-case criterion for safety, minimax policy optimization~\cite{li2019robust, zhang2019policy} or its generalization Stackelberg games~\cite{yang2022stackelberg, zhou2021decentralized, lauffer2022no, bai2021sample} are often the frameworks to promote resilience. Other works include meta-adaptive nonlinear control integrating learning modules for fast adaptation in unpredictable settings~\cite{shi2021meta, o2022neural}.

\section{Conclusions and Future Work} \label{Conclusions}
We proposed a novel constrained meta-Reinforcement Learning (RL) framework for adaptable safety guarantees in non-stationary environments. End-to-end differentiation is enabled via the differentiable convex programming, and the theoretical and empirical analysis demonstrated the advantages of our approach over benchmarks. We suggest future work that concentrates on enhancing the effectiveness and efficiency of generalizable AI, specifically by incorporating causality in scenarios with constraints. While meta-learning aims to leverage memory or training across multiple tasks for generalization, the incorporation of causality, which captures cause-and-effect relationships, has the potential to efficiently transfer knowledge from a particular task to different ones by revealing hidden environmental dynamics. Thus, fusing causality into the existing RL paradigm presents a promising avenue for more efficient learning and improved generalizability. Consequently, our future work will explore this direction to further enhance RL capabilities.




\bibliography{reference}

\end{document}